\documentclass[letterpaper, 10 pt, conference]{ieeeconf}

\IEEEoverridecommandlockouts                              

\usepackage{graphicx} 

\usepackage[noadjust]{cite}
\usepackage{hyperref}

\usepackage{amsfonts}

\usepackage[cmex10]{amsmath}
\usepackage{bm}

\DeclareMathOperator*{\argmin}{argmin}

\author{Ra\'ul Mur-Artal and Juan D. Tard\'os
\thanks{This work was supported by the Spanish government under Project DPI2015-67275, the Arag\'on regional governmnet under Project DGA T04-FSE and the Ministerio de Educaci\'on Scholarship FPU13/04175.}
\thanks{The authors are with the Instituto de Investigaci\'on en Ingenier\'ia de Arag\'on (I3A), Universidad de Zaragoza, Mar\'ia de Luna 1, 50018 Zaragoza, Spain. Email: 
        {\tt\small \{raulmur,tardos\}@unizar.es}.}
}

\title{\LARGE \bf Visual-Inertial Monocular SLAM with Map Reuse}

\begin{document}

\thispagestyle{empty} 
\onecolumn 
 
\begin{center} 
\noindent 
 
This paper has been accepted for publication in \emph{IEEE Robotics and Automation Letters}. 
 
\vspace{2em} 
 
DOI: \href{https://doi.org/10.1109/LRA.2017.2653359}{10.1109/LRA.2017.2653359} 
 
IEEE Xplore: \url{http://ieeexplore.ieee.org/document/7817784/} 
\end{center} 
\vspace{3em} 
 
\copyright2017 IEEE. 
Personal use of this material is permitted. Permission  
from IEEE must be obtained for all other uses, in any current or  
future media, including reprinting 
/republishing this material for  
advertising or promotional purposes, 
creating new collective works, for resale or  
redistribution to servers or lists, or reuse of any copyrighted  
component of this work in other works. 
 
\twocolumn

\setcounter{page}{1}

\maketitle
\thispagestyle{empty}
\pagestyle{empty}

\begin{abstract}

In recent years there have been excellent results in Visual-Inertial Odometry techniques, which aim to compute the incremental motion of the sensor with high accuracy and robustness. 
However these approaches lack the capability to close loops, and trajectory estimation accumulates drift even if the sensor is continually revisiting the same place. 
In this work we present a novel tightly-coupled Visual-Inertial Simultaneous Localization and Mapping system that is able to close loops and reuse its map to achieve zero-drift localization in 
already mapped areas. While our approach can be applied to any camera configuration, we address here the most general problem of a monocular camera, with its well-known
scale ambiguity. We also propose a novel IMU initialization method, which computes the scale, the gravity direction, the velocity, 
and gyroscope and accelerometer biases, in a few seconds with high accuracy. We test our system in the 11 sequences of a recent micro-aerial vehicle public dataset achieving a 
typical scale factor error of $1\%$ and centimeter precision. We compare to the state-of-the-art in visual-inertial odometry in sequences with revisiting,
proving the better accuracy of our method due to map reuse and no drift accumulation.

\end{abstract}


\section{INTRODUCTION}
Motion estimation from onboard sensors is currently a hot topic in Robotics and Computer Vision communities, as it can enable emerging technologies such as autonomous cars,
augmented and virtual reality, service robots and drone navigation. Among different sensor modalities, visual-inertial setups provide a cheap solution with great potential.
On the one hand, cameras provide rich information of the environment, which allows to build 3D models, localize the camera and recognize already visited places. On the other hand, IMU sensors
provide self-motion information, allowing to recover metric scale
 for monocular vision, and to estimate gravity direction, rendering absolute pitch and roll observable.

Visual-inertial fusion has been a very active research topic in the last years.
The recent research is focused on tightly-coupled (i.e. joint optimization of all sensor states) visual-inertial odometry, 
using filtering \cite{MourikisICRA07,WuRSS15,BloeschIROS15} or keyframe-based non-linear optimization \cite{indelmanRAS13,LeuteneggerIJRR14,EngelIMU,ForsterRSS15,AlejoICRA16}.
Nevertheless these approaches are only able to compute incremental motion and lack the capability to close loops and reuse a map of an already mapped environment. 
This implies that estimated trajectory 
accumulates drift without bound, even if the sensor is always moving in the same environment. This is due to the marginalization of past states  to maintain a constant computational cost \cite{MourikisICRA07,LeuteneggerIJRR14,WuRSS15,BloeschIROS15,EngelIMU}, or 
the use of full smoothing \cite{indelmanRAS13,ForsterRSS15}, with an almost constant complexity in exploration but that can be as expensive as a batch method in the presence of
loop closures. The filtering method \cite{JonesIJRR11}, is able to close loops topologically and 
reuse its map, but global metric consistency is not enforced in real-time. The recent system \cite{LynenRSS15} is able to reuse a given map, built offline, and perform
visual-inertial tracking.

\begin{figure}[t]
      \centering
      \includegraphics[width=0.95\linewidth]{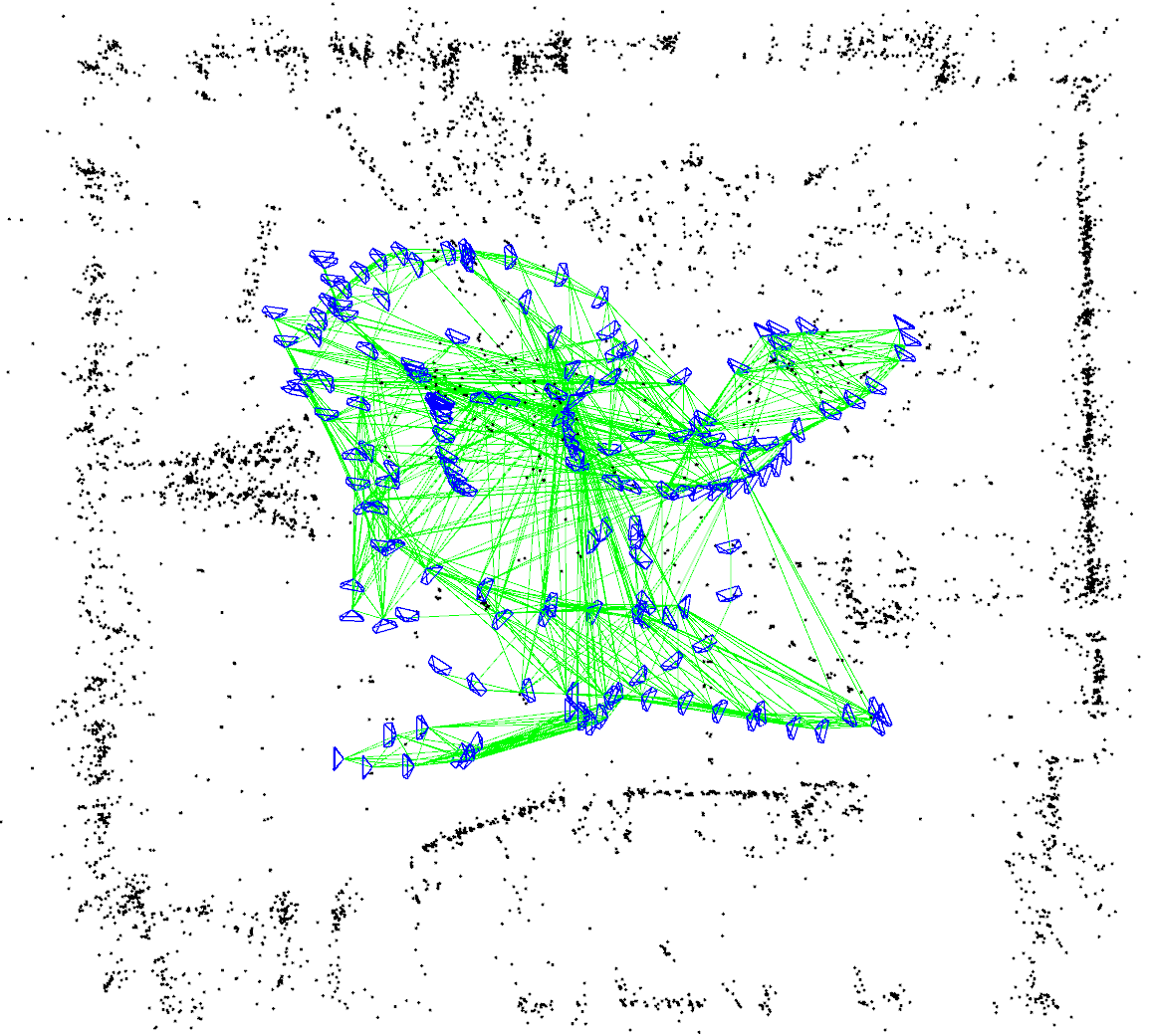}
      \caption{Estimated map and keyframes by our Visual-Inertial ORB-SLAM in \emph{V1\_02\_medium} from the EuRoC dataset \cite{EUROC}. The top view was rendered
      using the estimated gravity direction. The green lines connect keyframes that share more than 100 point observations
      and are a proof of the capability of the system to reuse the map, which, in contrast to visual-inertial odometry, allows zero-drift localization
      when continually revisiting the same place.}
      \label{fig:view}
   \end{figure}

Building on the preintegration of Lupton and Sukkarieh \cite{LuptonTRO12}, its application to the SO(3) manifold by Forster et al. \cite{ForsterRSS15}, and its
factor graph representation by Indelman et al. \cite{indelmanRAS13}, we present in this paper Visual-Inertial ORB-SLAM, to the best of our knowledge the 
first keyframe-based Visual-Inertial SLAM that is able to metrically close loops in real-time
and reuse the map that is being built online. 
Following the approach of ORB-SLAM \cite{MurTRO15}, inspired by the work of Klein and Murray \cite{ptam}, our tracking
optimizes the current frame assuming a fixed map, and our backend performs local Bundle Adjustment (BA), optimizing a local window of
keyframes, including an outer window of fixed keyframes. In contrast to full smoothing, this approach allows constant time local BA, and by
not marginalizing 
past states, we are able to reuse them. We detect large loops using place recognition and correct them using a lightweight pose-graph optimization, 
followed by full BA in a separate thread, not
to interfere with real-time operation. Fig. \ref{fig:view} shows a reconstruction by our system in a sequence with continuous revisiting.

Both our tracking and local BA fix states in their optimizations, which could potentially bias the solution. For this reason we need a reliable visual-inertial initialization
that provides accurate state estimations before we start fixing states. To this end we propose to perform a visual-inertial full BA that provides the optimal solution 
for structure, camera poses, scale, velocities, gravity, and gyroscope and accelerometer biases. This full BA is a non-linear optimization that requires 
a good initial seed to converge. We propose in Section \ref{sec:ini} a divide and conquer approach to compute this initial solution. We firstly process a few seconds of video 
with our pure monocular ORB-SLAM \cite{MurTRO15} to estimate an initial solution for structure and several keyframe poses, up to an unknown scale factor.
We then compute the bias of the gyroscope, which can be easily estimated from the known orientation of the keyframes, so that we can correctly rotate 
the accelerometer measurements. Then we solve scale and gravity without considering the accelerometer bias, using an approach inspired in \cite{LuptonTRO12}. 
To facilitate distinguishing between gravity and accelerometer bias, we use the knowledge of the magnitude of the gravity 
and solve for accelerometer bias, refining scale and gravity direction. At this point it is straightforward to retrieve the velocities for all keyframes.
Our experiments validate that this is an efficient, reliable and accurate initialization method. Moreover it is general, could be applied to any keyframe-based monocular SLAM, 
does not assume any initial condition, and just require a movement of the sensor that make all variables observable \cite{MartinelliIJCV14}.
While previous approaches \cite{MartinelliIJCV14, yangTASE16, KaiserRAL17} jointly solve vision and IMU, either ignoring gyroscope or accelerometer biases, we
efficiently compute all variables by subdividing the problem in simpler steps.
   
\section{VISUAL-INERTIAL PRELIMINARIES}

The input for our Visual-Inertial ORB-SLAM is a stream of IMU measurements and monocular camera frames. We consider a conventional 
pinhole-camera model \cite{Hartley} with a projection function $\pi : \mathbb{R}^3 \rightarrow \Omega$, which transforms 3D points 
$\mathbf{X_\mathtt{C}}\in\mathbb{R}^3$ in camera reference $\mathtt{C}$, into 2D points on the image plane $\mathbf{x_\mathtt{C}}\in\Omega \subset \mathbb{R}^2$:
\begin{equation}\label{eq:proj}
 \pi(\mathbf{X_\mathtt{C}}) = \begin{bmatrix} f_u \frac{X_\mathtt{C}}{Z_\mathtt{C}} + c_u \\[0.5em] f_v\frac{Y_\mathtt{C}}{Z_\mathtt{C}} + c_v \end{bmatrix},
 \quad \mathbf{X_\mathtt{C}}=\left[X_\mathtt{C}\,\,Y_\mathtt{C}\,\,Z_\mathtt{C}\right]^T
\end{equation}
where $\left[f_u\,\,f_v\right]^T$ is the focal length and $\left[c_u\,\,c_v\right]^T$ the principal point. This projection function does not consider the distortion produced by the 
camera lens. When we extract keypoints on the image, we undistort their coordinates so that they can be matched to projected points using \eqref{eq:proj}.

The IMU, whose reference we denote with $\mathtt{B}$, measures the acceleration $\mathbf{a}_\mathtt{B}$ and angular velocity $\bm{\omega_\mathtt{B}}$ 
of the sensor at regular intervals $\Delta t$, typically at hundreds of Hertzs. Both measurements are affected, in addition to sensor noise, by slowly varying biases $\mathbf{b}_a$ and $\mathbf{b}_g$ of the accelerometer and 
gyroscope respectively. Moreover the accelerometer is subject to gravity $\mathbf{g}_\mathtt{W}$ and one needs to subtract its effect to compute the motion. 
The discrete evolution of the IMU orientation $\mathbf{R_\mathtt{WB}} \in \mathrm{SO}(3)$, 
position $_\mathtt{W}\mathbf{p_\mathtt{B}}$ and velocity $_\mathtt{W}\mathbf{v_\mathtt{B}}$, in the world reference $\mathtt{W}$, can be computed as follows \cite{ForsterRSS15}:
\begin{equation}\label{eq:imu}
\begin{aligned}
 \mathbf{R}^{k+1}_\mathtt{WB} & =  \mathbf{R}^{k}_\mathtt{WB} \, \mathrm{Exp}\left(\left(\bm{\omega}^k_\mathtt{B} - \bm{b}^k_g\right)\Delta t\right) \\
 _\mathtt{W}\mathbf{v}^{k+1}_\mathtt{B} & =  {_\mathtt{W}\mathbf{v}^{k}_\mathtt{B}} + \mathbf{g}_\mathtt{W} \Delta t + \mathbf{R}^{k}_\mathtt{WB} \left(\bm{a}^k_\mathtt{B} - \bm{b}^k_a\right)\Delta t \\
 _\mathtt{W}\mathbf{p}^{k+1}_\mathtt{B} & =  {_\mathtt{W}\mathbf{p}^{k}_\mathtt{B}} + {_\mathtt{W}\mathbf{v}^{k}_\mathtt{B}} \Delta t + \frac{1}{2}\mathbf{g}_\mathtt{W} \Delta t^2 + \frac{1}{2} \mathbf{R}^{k}_\mathtt{WB} \left(\bm{a}^k_\mathtt{B} - \bm{b}^k_a\right)\Delta t^2
\end{aligned}
 \end{equation}

The motion between two consecutive keyframes can be defined in terms of the preintegration $\Delta \mathbf{R}$, $\Delta \mathbf{v}$ and $\Delta \mathbf{p}$ from all measurements 
in-between \cite{LuptonTRO12}. We use the recent IMU preintegration described in \cite{ForsterRSS15}:
\begin{equation}\label{eq:preint}
\begin{aligned}
\mathbf{R}^{i+1}_\mathtt{WB} & =  \mathbf{R}^{i}_\mathtt{WB}  \Delta \mathbf{R}_{i,i+1} \mathrm{Exp}\left(\left(\mathbf{J}^g_{\Delta R}\mathbf{b}^i_g\right)\right) \\
 _\mathtt{W}\mathbf{v}^{i+1}_\mathtt{B} & =  {_\mathtt{W}\mathbf{v}^{i}_\mathtt{B}} + \mathbf{g}_\mathtt{W} \Delta t_{i,i+1} \\ &+  \mathbf{R}^{i}_\mathtt{WB} \left( \Delta \mathbf{v}_{i,i+1} + \mathbf{J}^g_{\Delta v} \mathbf{b}^i_g  +  \mathbf{J}^a_{\Delta v} \mathbf{b}^i_a\right) \\
 _\mathtt{W}\mathbf{p}^{i+1}_\mathtt{B} & =  {_\mathtt{W}\mathbf{p}^{i}_\mathtt{B}} + {_\mathtt{W}\mathbf{v}^{i}_\mathtt{B}} \Delta t_{i,i+1} + \frac{1}{2}\mathbf{g}_\mathtt{W} \Delta t^2_{i,i+1} \\ & + \mathbf{R}^{i}_\mathtt{WB} \left(\Delta \mathbf{p}_{i,i+1} + \mathbf{J}^g_{\Delta p} \mathbf{b}^i_g +  \mathbf{J}^a_{\Delta p} \mathbf{b}^i_a\right)
\end{aligned}
 \end{equation}
where the Jacobians $\mathbf{J}^a_{(\cdot)}$ and  $\mathbf{J}^g_{(\cdot)}$ account for a first-order approximation of the effect of changing the biases without 
explicitly recomputing the preintegrations. Both preintegrations and Jacobians can be efficiently computed iteratively as IMU measurements arrive \cite{ForsterRSS15}.

Camera and IMU are considered rigidly attached and the transformation $\mathbf{T}_\mathtt{CB} = \left[\mathbf{R}_{\mathtt{CB}} | {_\mathtt{C}\mathbf{p}_\mathtt{B}}\right]$ between
their reference systems known from calibration \cite{kalibr}.

\section{VISUAL-INERTIAL ORB-SLAM}

The base of our visual-inertial system  is ORB-SLAM \cite{MurTRO15,orb2}. This system has three parallel threads for Tracking, Local Mapping and Loop Closing.
The system is designed to work on large scale environments, by building a covisibility graph that allows to recover local maps for tracking and mapping, and by performing lightweight pose-graph 
optimizations at loop closure. In addition ORB-SLAM allows to build a map of an environment and switch to a less CPU-intensive \emph{localization-only} 
mode (i.e. mapping and loop closing are disabled), thanks to the relocalization capability of the system. 
ORB-SLAM is open-source\footnote{\url{https://github.com/raulmur/ORB_SLAM2}} and has been extensively 
evaluated on public datasets achieving top performing results. 
In this section we detail the main changes in the Tracking, Local Mapping and Loop Closing threads with respect to the original system. The 
visual-inertial initialization is presented in Section \ref{sec:ini}.

\subsection{Tracking}

\newcommand{\scaletr}{0.15}
\begin{figure*}[t]
      \centering
      \includegraphics[width=\linewidth]{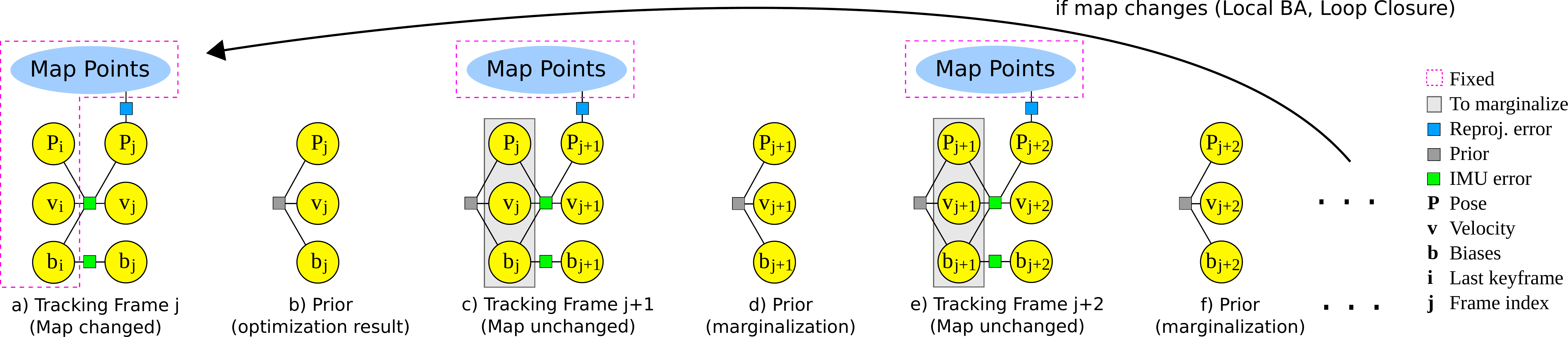} 
    
      \caption{Evolution of the optimization in the Tracking thread. (a) We start optimizing the frame $j$ linked by an IMU constraint to last keyframe $i$. 
      (b) The result of the optimization (estimation and Hessian matrix) serves as prior for next optimization. (c) When tracking next frame $j+1$,  
      both frames $j$ and $j+1$ are jointly optimized, being linked
      by an IMU constraint, and having frame $j$ the prior from previous optimization. (d) At the end of the optimization, the frame $j$ is marginalized out and the 
      result serves as prior for following optimization. (e-f) This process is repeated until there is a map update from the Local Mapping or Loop Closing thread. 
      In such case the optimization links the current frame to last keyframe discarding the prior, which is not valid after the map change.}
      \label{fig:tr}
   \end{figure*}

Our visual-inertial tracking is in charge of tracking the sensor pose, velocity and IMU biases, at frame-rate. This allows us to predict the camera pose 
very reliably, instead of using an ad-hoc motion model as in the original monocular system. Once the camera pose is predicted, the map points in the local map are 
projected and matched with keypoints on the frame. We then optimize current frame $j$ by minimizing the feature reprojection error of all matched 
points and an IMU error term. This optimization is different depending on the map being updated or not by the Local Mapping or the Loop Closing
thread, as illustrated in Fig. \ref{fig:tr}.

When tracking is performed just after a map update (Fig. \ref{fig:tr}a) the IMU error term links current frame j with last keyframe $i$:
\begin{equation}\label{eq:tr1}
\begin{gathered}
 \theta = \left\{\mathbf{R}^j_\mathtt{WB},{_\mathtt{W}\mathbf{p}^j_\mathtt{B}},{_\mathtt{W}\mathbf{v}^j_\mathtt{B}}, \mathbf{b}^j_g, \mathbf{b}^j_a\right\} \\
 \theta^* = 
 \argmin_{\theta} \left(
 \sum_k \mathbf{E}_\mathrm{proj} (k,j) + \mathbf{E}_\mathrm{IMU} (i,j)\right)
 \end{gathered}
\end{equation}
where the feature reprojection error $\mathbf{E}_\mathrm{proj}$ for a given match $k$, is defined as follows:
\begin{equation}\label{eq:Ereproj}
\begin{gathered}
 \mathbf{E}_\mathrm{proj}(k,j) = \rho \left(\left(\mathbf{x}^k - \pi(\mathbf{X}^k_\mathtt{C})\right)^T \bm{\Sigma_k} \left( \mathbf{x}^k - \pi(\mathbf{X}^k_\mathtt{C})\right) \right) \\
 \mathbf{X}^k_\mathtt{C} = \mathbf{R}_\mathtt{CB} \mathbf{R}^j_\mathtt{BW} \left(\mathbf{X}^k_\mathtt{W}- {_\mathtt{W}\mathbf{p}^j_\mathtt{B}}\right) + {_\mathtt{C}\mathbf{p}_\mathtt{B}}
 \end{gathered}
\end{equation}
where $\mathbf{x}^k$ is the keypoint location in the image, $\mathbf{X}^k_\mathtt{W}$ the map point in world coordinates, and $\bm{\Sigma_k}$ 
the information matrix associated to the keypoint scale. The IMU error term $\mathbf{E}_\mathrm{IMU}$ is:
\begin{equation}\label{eq:Eimu}
\begin{gathered}
 \mathbf{E}_\mathrm{IMU} (i,j)  = \rho \left(\left[\mathbf{e}^T_R \, \mathbf{e}^T_v \, \mathbf{e}^T_p\right]
 \bm{\Sigma}_I \left[\mathbf{e}^T_R \, \mathbf{e}^T_v \, \mathbf{e}^T_p\right]^T\right) \\ + \rho \left(\mathbf{e}^T_b\bm{\Sigma}_{R}\mathbf{e}_b\right)
\\
\begin{aligned}
 \mathbf{e}_R & = \mathrm{Log} \left(\left(\Delta \mathbf{R}_{ij} \mathrm{Exp}\left(\mathbf{J}^g_{\Delta R}\mathbf{b}^j_g \right)\right)^T
 \mathbf{R}_\mathtt{BW}^i \mathbf{R}_\mathtt{WB}^j\right)
\\
\mathbf{e}_v & = \mathbf{R}_\mathtt{BW}^i \left(_\mathtt{W}\mathbf{v}_\mathtt{B}^j-{_\mathtt{W}\mathbf{v}_\mathtt{B}^i}- \mathbf{g}_\mathtt{W} \Delta t_{ij}\right)
\\ & -\left( \Delta \mathbf{v}_{ij} + \mathbf{J}^g_{\Delta v} \mathbf{b}^j_g  +  \mathbf{J}^a_{\Delta v} \mathbf{b}^j_a\right)
\\
\mathbf{e}_p & = \mathbf{R}_\mathtt{BW}^i \left(_\mathtt{W}\mathbf{p}_\mathtt{B}^j-{_\mathtt{W}\mathbf{p}_\mathtt{B}^i} -{_\mathtt{W}\mathbf{v}_\mathtt{B}^i} \Delta t_{ij} - \frac{1}{2} \mathbf{g}_\mathtt{W} \Delta t_{ij}^2\right)
\\ & -\left( \Delta \mathbf{p}_{ij} + \mathbf{J}^g_{\Delta p} \mathbf{b}^j_g  +  \mathbf{J}^a_{\Delta p} \mathbf{b}^j_a\right)
\\
\mathbf{e}_b & = \mathbf{b}^j-\mathbf{b}^i
 \end{aligned}
 \end{gathered}
 \end{equation}
 where $\bm{\Sigma}_I$ is the information matrix of the preintegration and $\bm{\Sigma}_R$ of the bias random walk  \cite{ForsterRSS15}, and $\rho$ is the Huber robust cost function. We solve this optimization problem with Gauss-Newton algorithm implemented
 in g2o \cite{g2o}. After the optimization (Fig. \ref{fig:tr}b) the resulting estimation and Hessian matrix serves as prior for next optimization. 
 
 Assuming no map update (Fig. \ref{fig:tr}c), the next frame $j+1$ will be optimized with a link to frame $j$ and using the prior 
 computed at the end of the previous optimization (Fig \ref{fig:tr}b):
\begin{equation}\label{eq:tr2}
\begin{gathered}
 \theta = \left\{\mathbf{R}^j_\mathtt{WB},\mathbf{p}^j_\mathtt{W},\mathbf{v}^j_\mathtt{W}, \mathbf{b}^j_g, \mathbf{b}^j_a,
 \mathbf{R}^{j+1}_\mathtt{WB},\mathbf{p}^{j+1}_\mathtt{W},\mathbf{v}^{j+1}_\mathtt{W}, \mathbf{b}^{j+1}_g, \mathbf{b}^{j+1}_a\right\} \\
 \theta^* = 
 \argmin_{\theta}
 \Big(\sum_k \mathbf{E}_\mathrm{proj} (k,j+1)  + \mathbf{E}_\mathrm{IMU} (j,j+1) \\ + \mathbf{E}_\mathrm{prior} (j)\Big)
 \end{gathered}
\end{equation}
where $\mathbf{E}_\mathrm{prior}$ is a prior term:
\begin{equation}
\begin{gathered}
 \mathbf{E}_\mathrm{prior} (j)  = \rho \left(\left[\mathbf{e}^T_R \, \mathbf{e}^T_v \, \mathbf{e}^T_p \, \mathbf{e}^T_b\right] \bm{\Sigma}_p 
 \left[\mathbf{e}^T_R \, \mathbf{e}^T_v \, \mathbf{e}^T_p \, \mathbf{e}^T_b\right]^T\right)
\\
\begin{aligned}
 \mathbf{e}_R &= \mathrm{Log} \left(\mathbf{\bar{R}}_\mathtt{BW}^j \mathbf{R}_\mathtt{WB}^j\right)
&
\mathbf{e}_v  &= \mathbf{_\mathtt{W}\bar{v}}_\mathtt{B}^j-{_\mathtt{W}\mathbf{v}_\mathtt{B}^j }
\\
\mathbf{e}_p  &= {_\mathtt{W}\mathbf{\bar{p}}_\mathtt{B}^j}-{_\mathtt{W}\mathbf{p}_\mathtt{B}^j} 
&
\mathbf{e}_b  &= \mathbf{\bar{b}}^j-\mathbf{b}^j
\end{aligned}
\end{gathered}
 \end{equation}
 where $(\bar{\cdot})$ and $\bm{\Sigma}_{p}$ are the estimated states and Hessian matrix resulting from previous optimization (Fig. \ref{fig:tr}b).
 After this optimization (Fig. \ref{fig:tr}d), frame $j$ is marginalized out \cite{LeuteneggerIJRR14}. 
 This optimization linking two consecutive frames and using a prior is repeated (Fig. \ref{fig:tr}e-f) until a map change, when the 
 prior will be no longer valid and the tracking will link again the current frame to the last keyframe (Fig. \ref{fig:tr}a). Note that this 
 is the optimization, Fig \ref{fig:tr} (e-f), that is always performed in \emph{localization-only} mode, as the map is not updated.

\subsection{Local Mapping}\label{sec:mapping}

\begin{figure}[t]
      \centering
      \includegraphics[width=0.78\linewidth]{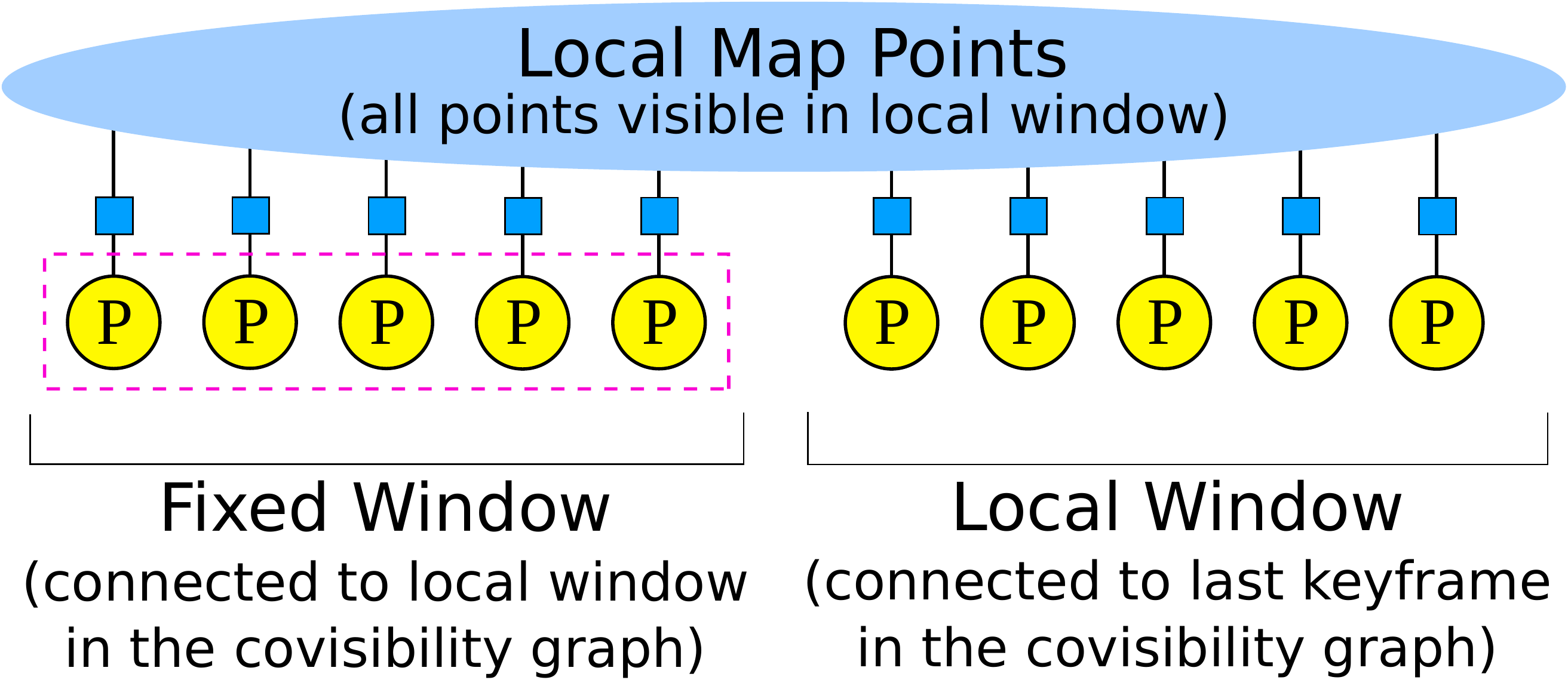}  \\[0.5em]
      \makebox[\linewidth][c]{ORB-SLAM's Local BA} \\[1.5em]
      \includegraphics[width=0.78\linewidth]{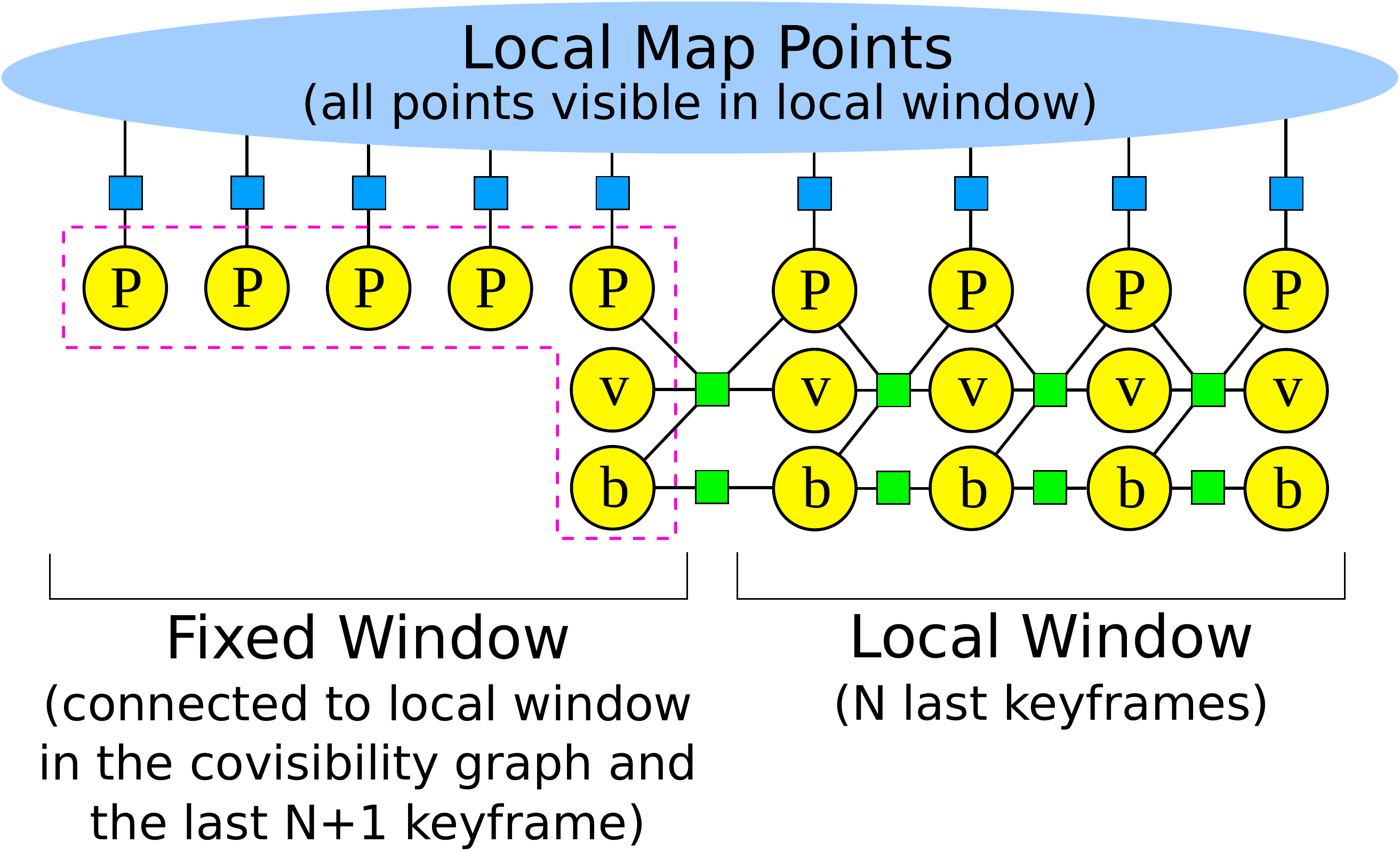}  \\[0.5em]
      \makebox[\linewidth][c]{Visual-Inertial ORB-SLAM's Local BA} 
      \caption{Comparison of Local Bundle Adjustment between original ORB-SLAM (top) and proposed Visual-Inertial ORB-SLAM (bottom). The local window in Visual-Inertial
      ORB-SLAM is retrieved by temporal order of keyframes, while in ORB-SLAM is retrieved using the covisibility graph.}
      \label{fig:ba}      
   \end{figure}

The Local Mapping thread performs local BA after a new keyframe insertion. It optimizes the last $N$ keyframes (local window) and all points seen by those $N$ keyframes. 
All other keyframes that share observations of local points (i.e. are connected in the covisibility graph to any local keyframe), but are not in the local window, 
contribute to the total cost but are fixed during optimization (fixed window). The keyframe $N+1$ is always included in the fixed window as it constrains the IMU states. 
Fig. \ref{fig:ba} illustrates the differences between local BA in original ORB-SLAM and Visual-Inertial ORB-SLAM. The cost function is a combination of IMU error 
terms \eqref{eq:Eimu} and 
reprojection error terms \eqref{eq:Ereproj}. Note that the visual-inertial version, compared to
the vision only, is more complex as there are 9 additional states (velocity and biases) to optimize per keyframe. A suitable local window size has to be chosen for real-time 
performance.

The Local Mapping is also in charge of keyframe management. The original ORB-SLAM policy discards redundant keyframes,
so that map size does not grow if localizing in a well mapped area. This policy is problematic when using IMU information, which constrains the motion of consecutive 
keyframes. The longer the temporal difference between consecutive keyframes, the weaker information IMU provides. Therefore we allow the mapping to discard 
redundant keyframes, if that does not make two consecutive keyframes in the local window of local BA to differ more than $0.5\mathrm{s}$. To be able to perform full BA, after a 
loop closure or 
at any time to refine a map, we do not allow any two consecutive keyframes to differ more than 3s. If we switched-off full BA with IMU constraints, we would only need to restrict
the temporal offset between keyframes in the local window.

\subsection{Loop Closing}

The loop closing thread aims to reduce the drift accumulated during exploration, when returning to an already mapped area. The place recognition module matches
a recent keyframe with a past keyframe. This match is validated computing a rigid body transformation that aligns matched points between keyframes \cite{horn}. Finally
an optimization is carried out to reduce the error accumulated in the trajectory. This optimization might be very costly in large maps, therefore the strategy
is to perform a pose-graph optimization, which reduces the complexity, as structure is ignored, and exhibits good convergence as shown in \cite{MurTRO15}.
In contrast to the original ORB-SLAM, we perform the pose-graph optimization on 6 Degrees of Freedom( DoF) instead of 7 DoF \cite{haukeScale}, as scale is observable.
This pose-graph ignores IMU information, not optimizing velocity or IMU biases. 
Therefore we correct velocities by rotating them according to the 
corrected orientation of the associated keyframe. While this is not optimal, biases and velocities should be locally accurate to continue using IMU information right after 
pose-graph optimization. We perform afterwards a full BA in a parallel thread that optimizes all states, including velocities and biases.

\section{IMU INITIALIZATION}\label{sec:ini}

We propose in this section a method to compute an initial estimation for a visual-inertial full BA of the scale, gravity direction, velocity and IMU biases, given a set of
keyframes processed by a monocular SLAM algorithm.
The idea is to run the monocular SLAM for a few seconds, assuming the sensor performs a motion that makes all variables
observable. While we build on ORB-SLAM \cite{MurTRO15}, any other SLAM could be used. The only requirement is that any two consecutive keyframes are close in time (see Section \ref{sec:mapping}), 
to reduce IMU noise integration.

The initialization is divided in simpler subproblems: (1) gyroscope bias estimation, (2) scale and gravity approximation, considering no accelerometer
bias, (3) accelerometer bias estimation, and scale and gravity direction refinement, and (4) velocity estimation. 

\subsection{Gyroscope Bias Estimation}
Gyroscope bias can be estimated just from the known orientation of two consecutive keyframes. Assuming a negligible bias change, we optimize a
constant bias $\mathbf{b}_g$, which minimizes the difference between gyroscope integration and relative orientation computed from ORB-SLAM,
for all pairs of consecutive keyframes:
\begin{equation}\label{eq:gyro}
 \argmin_{ \mathbf{b}_g} \sum_{i=1}^{N-1}\left\|\mathrm{Log}\left(\left(\Delta \mathbf{R}_{i,i+1} \mathrm{Exp}\left(\mathbf{J}^g_{\Delta R}\mathbf{b}_g\right)\right)^T
 \mathbf{R}_\mathtt{BW}^{i+1} \mathbf{R}_\mathtt{WB}^i\right)\right\|^2
 \end{equation}
 where $N$ is the number of keyframes. $\mathbf{R}_\mathtt{WB}^{(\cdot)} = \mathbf{R}_\mathtt{WC}^{(\cdot)} \mathbf{R}_\mathtt{CB}$ 
 is computed from the orientation $\mathbf{R}_\mathtt{WC}^{(\cdot)}$ computed by ORB-SLAM and calibration $\mathbf{R}_\mathtt{CB}$. $\Delta \mathbf{R}_{i,i+1}$ is the 
 gyroscope integration between two consecutive keyframes. We solve \eqref{eq:gyro} with Gauss-Newton with a zero bias seed. Analytic jacobians for a similar expression
 can be found in \cite{ForsterRSS15}. 

\subsection{Scale and Gravity Approximation (no accelerometer bias)}
Once we have estimated the gyroscope bias, we can preintegrate velocities and positions, rotating correctly the acceleration measurements compensating the gyroscope bias.

The scale of the camera trajectory computed by ORB-SLAM is arbitrary. 
Therefore we need to include a scale factor $s$ when transforming between camera $\mathtt{C}$ and IMU $\mathtt{B}$ coordinate systems:
\begin{equation}\label{eq:s}
 {_\mathtt{W}\mathbf{p}_\mathtt{B}} = s \, {_\mathtt{W}\mathbf{p}_\mathtt{C}} +  \mathbf{R}_\mathtt{WC} \, {_\mathtt{C}\mathbf{p}_\mathtt{B}}
\end{equation}

Substituting \eqref{eq:s} into the equation relating position of two consecutive keyframes \eqref{eq:preint}, 
and neglecting at this point accelerometer bias, it follows:
\begin{equation}\label{eq:pnobias}
 \begin{aligned}
 s \, {_\mathtt{W}\mathbf{p}^{i+1}_\mathtt{C}}  & =  s \, {_\mathtt{W}\mathbf{p}^{i}_\mathtt{C}} + {_\mathtt{W}\mathbf{v}^{i}_\mathtt{B}} \Delta t_{i,i+1} + \frac{1}{2}\mathbf{g}_\mathtt{W} \Delta t^2_{i,i+1} 
 \\ 
 & + \mathbf{R}^{i}_\mathtt{WB} \Delta \mathbf{p}_{i,i+1} + \left(\mathbf{R}^{i}_\mathtt{WC}-\mathbf{R}^{i+1}_\mathtt{WC}\right){_\mathtt{C}\mathbf{p}_\mathtt{B}}
 \end{aligned}
\end{equation}

The goal is to estimate $s$ and $\mathbf{g}_\mathtt{W}$ by solving a linear system of equations on those variables. To avoid solving for $N$ velocities, and reduce complexity, 
we consider two relations \eqref{eq:pnobias} between three consecutive keyframes and use velocity relation in \eqref{eq:preint}, which results in the following expression:
\begin{equation}\label{eq:sys1}
\begin{bmatrix} \bm{\lambda}(i) & \bm{\beta}(i) \end{bmatrix}
\begin{bmatrix} s \\ \mathbf{g}_\mathtt{W} \end{bmatrix}
= \bm{\gamma}(i) 
\end{equation}
where, writing keyframes $i,i+1,i+2$ as $1,2,3$ for clarity of notation, we have:
\begin{equation}\label{eq:sys1-2}
\begin{aligned}
 \bm{\lambda}(i) & = \left({_\mathtt{W}\mathbf{p}^{2}_\mathtt{C}}-{_\mathtt{W}\mathbf{p}^{1}_\mathtt{C}}\right) \Delta t_{23}
 - \left({_\mathtt{W}\mathbf{p}^{3}_\mathtt{C}}-{_\mathtt{W}\mathbf{p}^{2}_\mathtt{C}}\right) \Delta t_{12}
 \\
 \bm{\beta}(i) & =  \frac{1}{2} \mathbf{I}_{3\times3}\left(\Delta t_{12}^2\Delta t_{23}+\Delta t_{23}^2\Delta t_{12}\right)
 \\
 \bm{\gamma}(i) & = \left(\mathbf{R}^{2}_\mathtt{WC}-\mathbf{R}^{1}_\mathtt{WC}\right){_\mathtt{C}\mathbf{p}_\mathtt{B}} \Delta t_{23}
 - \left(\mathbf{R}^{3}_\mathtt{WC}-\mathbf{R}^{2}_\mathtt{WC}\right){_\mathtt{C}\mathbf{p}_\mathtt{B}}\Delta t_{12} 
 \\
 & +  \mathbf{R}^{2}_\mathtt{WB} \Delta \mathbf{p}_{23}  \Delta t_{12} +  \mathbf{R}^{1}_\mathtt{WB} \Delta \mathbf{v}_{12} \Delta t_{12} \Delta t_{23}
 \\
 & - \mathbf{R}^{1}_\mathtt{WB} \Delta \mathbf{p}_{12} \Delta t_{23} 
 \end{aligned}
\end{equation}

We stack then all relations of three consecutive keyframes \eqref{eq:sys1} into a system $\mathbf{A}_{3(N-2)\times4} \, \mathbf{x}_{4\times1}=\mathbf{B}_{3(N-2)\times1}$ which can be solved via Singular Value 
Decomposition (SVD) to get the scale factor $s^*$ and gravity vector $\mathbf{g}^*_\mathtt{W}$. Note that we have $3(N-2)$ equations and 4 unknowns,
therefore we need at least 4 keyframes. 

\subsection{Accelerometer Bias Estimation, and Scale and Gravity Direction Refinement}

So far we have not considered accelerometer bias when computing scale and gravity. Just incorporating accelerometer biases in \eqref{eq:sys1}
will heavily increase the chance of having an ill-conditioned system, because gravity and accelerometer biases are hard to distinguish \cite{MartinelliIJCV14}.
To increase observability we introduce new information we did not consider so far, which is the gravity magnitude $G$. Consider an inertial reference $I$
with the gravity direction $\hat{\mathbf{g}}_\mathtt{I}=\{0,0,-1\}$, and the already computed gravity direction 
$\hat{\mathbf{g}}_\mathtt{W}={\mathbf{g}^*_\mathtt{W}}/{\|\mathbf{g}^*_\mathtt{W}\|}$. We can compute rotation $\mathbf{R}_\mathtt{WI}$ as follows:
\begin{equation}
\begin{gathered}
 \mathbf{R}_\mathtt{WI} = \mathrm{Exp(\mathbf{\hat{v}}\theta)} \\
 \mathbf{\hat{v}} = \frac{\hat{\mathbf{g}}_\mathtt{I} \times \hat{\mathbf{g}}_\mathtt{W}}{\|\hat{\mathbf{g}}_\mathtt{I} \times \hat{\mathbf{g}}_\mathtt{W}\|}
 , \quad \theta = \mathrm{atan2}\left(\|\hat{\mathbf{g}}_\mathtt{I} \times \hat{\mathbf{g}}_\mathtt{W}\|,\hat{\mathbf{g}}_\mathtt{I} \cdot \hat{\mathbf{g}}_\mathtt{W}\right)
 \end{gathered}
\end{equation}
and express now the gravity vector as:
\begin{equation}\label{eq:gr}
 \mathbf{g}_\mathtt{W} =  \mathbf{R}_\mathtt{WI} \, \hat{\mathbf{g}}_\mathtt{I} \, G
\end{equation}
where $\mathbf{R}_\mathtt{WI}$ can be parametrized with just two angles around x and y axes in $\mathtt{I}$, because a rotation around z axis, which is aligned with gravity, has no effect in
$\mathbf{g}_\mathtt{W}$. This rotation can be optimized using a perturbation $\bm{\delta\theta}$:
\begin{equation}\label{eq:gr2}
\begin{gathered}
 \mathbf{g}_\mathtt{W} =  \mathbf{R}_\mathtt{WI} \mathrm{Exp}(\bm{\delta \theta}) \, \hat{\mathbf{g}}_\mathtt{I} \, G
 \\
 \bm{\delta \theta} = \left[\bm{\delta \theta^T_{xy}} \,\, 0\right]^T, \quad \bm{\delta \theta_{xy}}= \left[\delta\theta_x  \,\, \delta\theta_y \right]^T
 \end{gathered}
\end{equation}
with a first-order approximation:
\begin{equation}\label{eq:gr3}
 \mathbf{g}_\mathtt{W} \approx  \mathbf{R}_\mathtt{WI} \, \hat{\mathbf{g}}_\mathtt{I} \, G -\mathbf{R}_\mathtt{WI} \, (\hat{\mathbf{g}}_\mathtt{I})_\times G \, \bm{\delta \theta} 
\end{equation}

Substituting \eqref{eq:gr3} in \eqref{eq:pnobias} and including now the effect of accelerometer bias, we obtain:
\begin{equation}\label{eq:pbias}
 \begin{aligned}
 s \, {_\mathtt{W}\mathbf{p}^{i+1}_\mathtt{C}}  & =  s \, {_\mathtt{W}\mathbf{p}^{i}_\mathtt{C}} + 
 {_\mathtt{W}\mathbf{v}^{i}_\mathtt{B}} \Delta t_{i,i+1} - \frac{1}{2} \mathbf{R}_\mathtt{WI} \, (\hat{\mathbf{g}}_\mathtt{I})_\times G \Delta t^2_{i,i+1} \,  \bm{\delta \theta}
 \\ 
 & + \mathbf{R}^{i}_\mathtt{WB} \left(\Delta \mathbf{p}_{i,i+1}+ \mathbf{J}^a_{\Delta p} \mathbf{b}_a\right) + \left(\mathbf{R}^{i}_\mathtt{WC}-\mathbf{R}^{i+1}_\mathtt{WC}\right){_\mathtt{C}\mathbf{p}_\mathtt{B}}
 \\
 & +\frac{1}{2} \mathbf{R}_\mathtt{WI} \, \hat{\mathbf{g}}_\mathtt{I} \, G \Delta t^2_{i,i+1}
 \end{aligned}
\end{equation}

Considering three consecutive keyframes as in \eqref{eq:sys1} we can eliminate velocities and get the following relation:
\begin{equation}\label{eq:sys2}
\begin{bmatrix} \bm{\lambda}(i) & \bm{\phi}(i) & \bm{\zeta}(i)\end{bmatrix}
\begin{bmatrix} s \\\bm{\delta \theta_{xy}}  \\ \mathbf{b}_a\end{bmatrix}
= \bm{\psi}(i) 
\end{equation}
where  $\bm{\lambda}(i)$ remains the same as in \eqref{eq:sys1-2}, and $\bm{\phi}(i)$, $\bm{\zeta}(i)$, and $\bm{\psi}(i) $ are computed as follows:
\begin{equation}\label{eq:sys2-2}
\begin{aligned}
 \bm{\phi}(i) & =  \left[\frac{1}{2} \mathbf{R}_\mathtt{WI} \, (\hat{\mathbf{g}}_\mathtt{I})_\times G\left(\Delta t_{12}^2\Delta t_{23}+\Delta t_{23}^2\Delta t_{12}\right)\right]_{(:,1:2)}
 \\
 \bm{\zeta}(i) & = \mathbf{R}^{2}_\mathtt{WB} \mathbf{J}^a_{\Delta p_{23}}\Delta t_{12} + \mathbf{R}^{1}_\mathtt{WB} \mathbf{J}^a_{\Delta v_{23}}\Delta t_{12}\Delta t_{23}
 \\
 & - \mathbf{R}^{1}_\mathtt{WB} \mathbf{J}^a_{\Delta p_{12}}\Delta t_{23}
 \\
 \bm{\psi}(i) & = \left(\mathbf{R}^{2}_\mathtt{WC}-\mathbf{R}^{1}_\mathtt{WC}\right){_\mathtt{C}\mathbf{p}_\mathtt{B}} \Delta t_{23}
 - \left(\mathbf{R}^{3}_\mathtt{WC}-\mathbf{R}^{2}_\mathtt{WC}\right){_\mathtt{C}\mathbf{p}_\mathtt{B}}\Delta t_{12} 
 \\
 & +  \mathbf{R}^{2}_\mathtt{WB} \Delta \mathbf{p}_{23}  \Delta t_{12} +  \mathbf{R}^{1}_\mathtt{WB} \Delta \mathbf{v}_{12} \Delta t_{12} \Delta t_{23}
 \\
 & - \mathbf{R}^{1}_\mathtt{WB} \Delta \mathbf{p}_{12} \Delta t_{23} +\frac{1}{2} \mathbf{R}_\mathtt{WI} \, \hat{\mathbf{g}}_\mathtt{I} \, G \Delta t^2_{ij}
 \end{aligned}
\end{equation}
where $[\,]_{(:,1:2)}$ means the first two columns of the matrix. 
Stacking all relations between three consecutive keyframes \eqref{eq:sys2} we form a linear system of equations $\mathbf{A}_{3(N-2)\times6} \, \mathbf{x}_{6\times1}=\mathbf{B}_{3(N-2)\times1}$ 
which can be solved via SVD to get the scale factor $s^*$, gravity direction correction $\bm{\delta \theta^*_{xy}} $ and accelerometer bias $\mathbf{b}^*_a$. 
In this case we have $3(N-2)$ equations and 6 unknowns and we need again at least 4 keyframes to solve the system. We can compute the condition number (i.e. the ratio between the 
maximum and minimum singular value) to check if the problem is well-conditioned (i.e. the sensor has performed a motion that makes all variables observable).
We could relinearize \eqref{eq:gr3} and iterate the solution, but in practice we found that a second iteration does not produce a noticeable improvement.

\subsection{Velocity Estimation}
We considered relations of three consecutive keyframes in equations \eqref{eq:sys1} and \eqref{eq:sys2}, so that the resulting linear systems do not have the $3N$ 
additional unknowns corresponding to velocities. The velocities for all keyframes can now be computed using equation \eqref{eq:pbias}, as scale, gravity and bias are known. To 
compute the velocity of the most recent keyframe, we use the velocity relation in \eqref{eq:preint}.

\subsection{Bias Reinitialization after Relocalization}
When the system relocalizes after a long period of time, using place recognition, we reinitialize gyroscope biases by solving \eqref{eq:gyro}. The accelerometer bias is estimated 
by solving a simplified \eqref{eq:sys2},  where the only unknown is the bias, as scale and gravity are already known. We use 20 consecutive frames localized with only vision to 
estimate both biases.

\section{EXPERIMENTS}
We evaluate the proposed IMU initialization method, detailed in Section \ref{sec:ini} and our Visual-Inertial ORB-SLAM in the EuRoC dataset \cite{EUROC}. 
It contains 11 sequences recorded from a micro aerial vehicle (MAV), flying around two different rooms and an industrial environment. Sequences are 
classified as \emph{easy}, \emph{medium} and \emph{difficult}, depending on illumination, texture, fast/slow motions or motion blur. 
The dataset provides synchronized global shutter WVGA stereo images at $20\mathrm{Hz}$ with IMU measurements at $200\mathrm{Hz}$ and trajectory ground-truth. These characteristics 
make it a 
really useful dataset to test Visual-Inertial SLAM systems. The 
experiments were performed processing left images only, in an Intel Core  i7-4700MQ computer with 8Gb  RAM.

\subsection{IMU Initialization}
We evaluate the IMU initialization in sequence \emph{V2\_01\_easy}. We run the initialization from scratch every time a keyframe is inserted by 
ORB-SLAM. We run the sequence at a lower frame-rate so that the repetitive initialization does not interfere with the normal behavior of the system. 
The goal is to analyze the convergence of the variables as more keyframes, i.e. longer trajectories, are processed by the initialization algorithm. Fig. \ref{fig:ini}
shows the estimated scale and IMU biases. It can be seen that between 10 and 15 seconds all variables have already 
converged to stable values and that the estimated scale factor is really close to its optimal value. This optimal scale factor is computed aligning the estimated trajectory 
with the ground-truth 
by a similarity transformation \cite{horn}. Fig. \ref{fig:ini} also shows the condition number of \eqref{eq:sys2}, indicating that some time is required 
to get a well-conditioned problem.
This confirms that the sensor has to perform a motion that makes all variables observable, especially the accelerometer bias. 
The last row in Fig. \ref{fig:ini} shows the time spent by the initialization algorithm, which exhibits a linear growth. 
This complexity is the result of not including velocities in \eqref{eq:sys1} and \eqref{eq:sys2}, which would have resulted in a 
quadratic complexity when using SVD to solve these systems. Subdividing the initialization in simpler subproblems, in contrast to \cite{MartinelliIJCV14,KaiserRAL17}, 
results in a very efficient method.

The proposed initialization allows to start fusing IMU information, as gravity, biases, scale and velocity are reliably estimated. For the EuRoC dataset, 
we observed that 15 seconds of MAV exploration give always an accurate initialization. As a future work we would like to investigate an automatic criterion 
to decide when we can consider an initialization successful, as we observed that an absolute threshold on the condition number is not reliable enough.

\newcommand{\scale}{0.83}
   \begin{figure}[t] 
      \centering
      \includegraphics[width=\scale\linewidth]{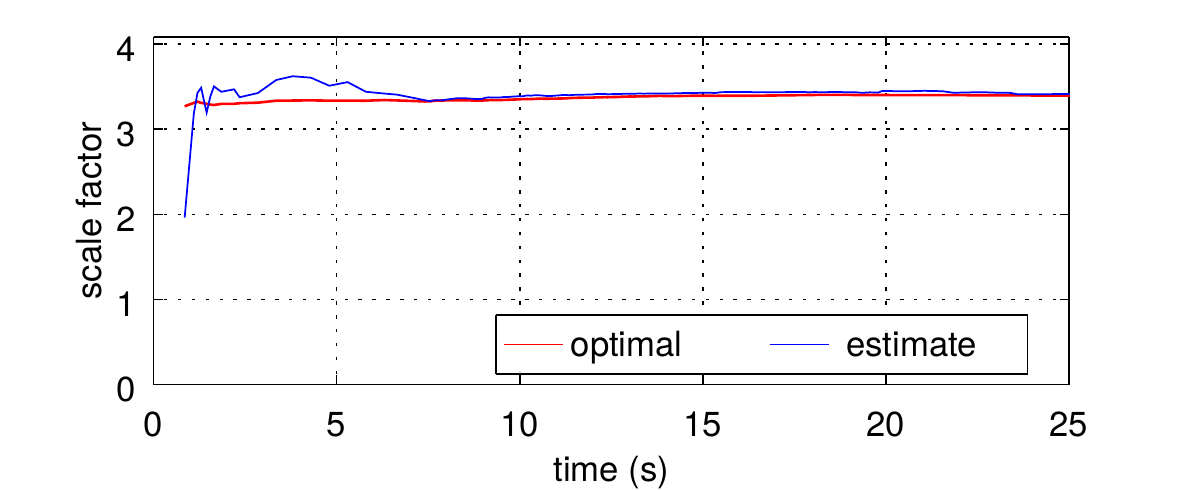}      
      \\
      \includegraphics[width=\scale\linewidth]{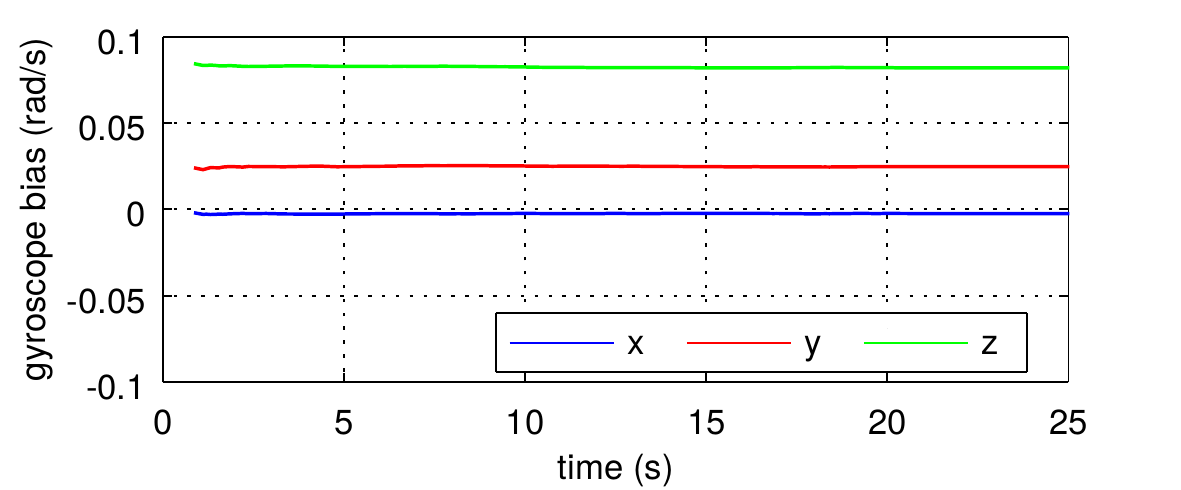}
      \\
      \includegraphics[width=\scale\linewidth]{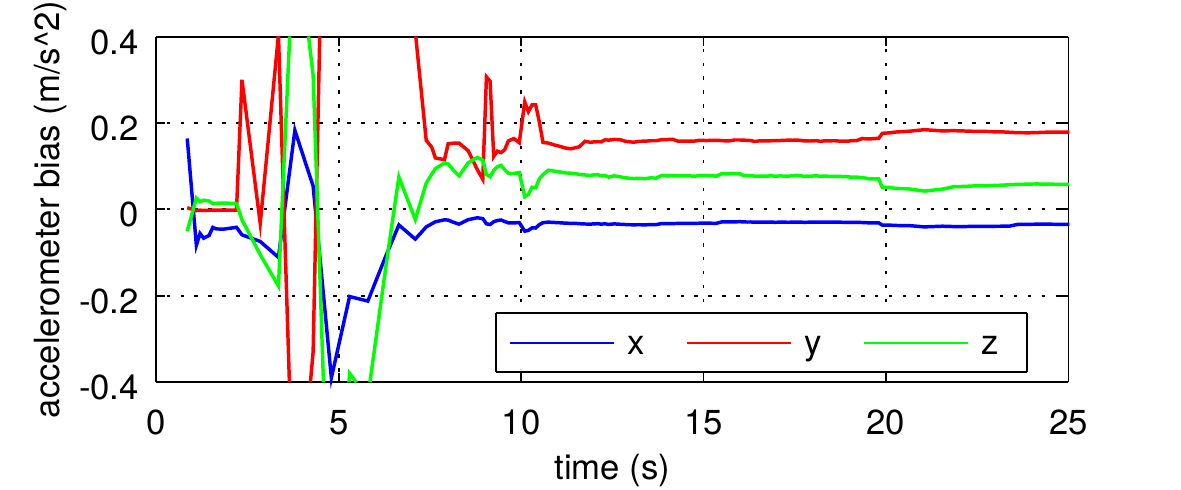}      
      \\
      \includegraphics[width=\scale\linewidth]{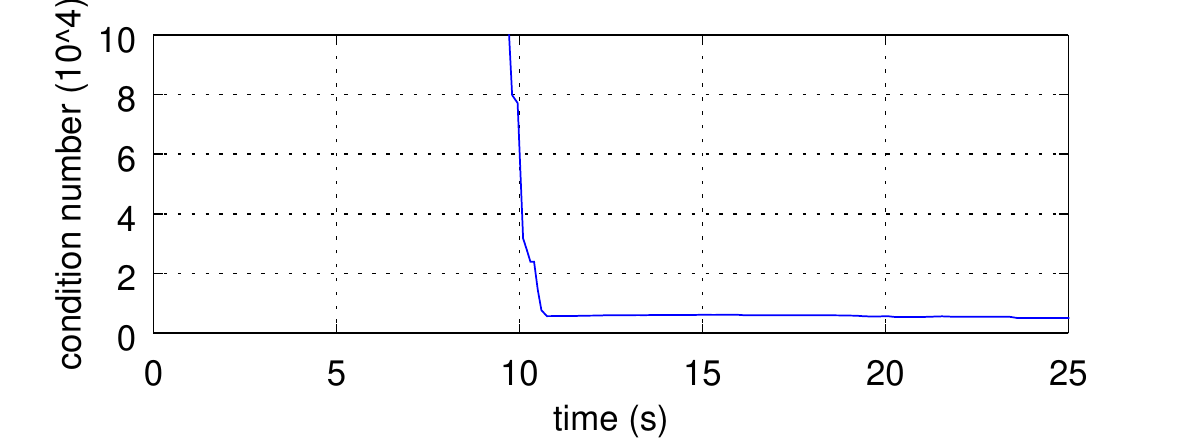}      
      \\
      \includegraphics[width=\scale\linewidth]{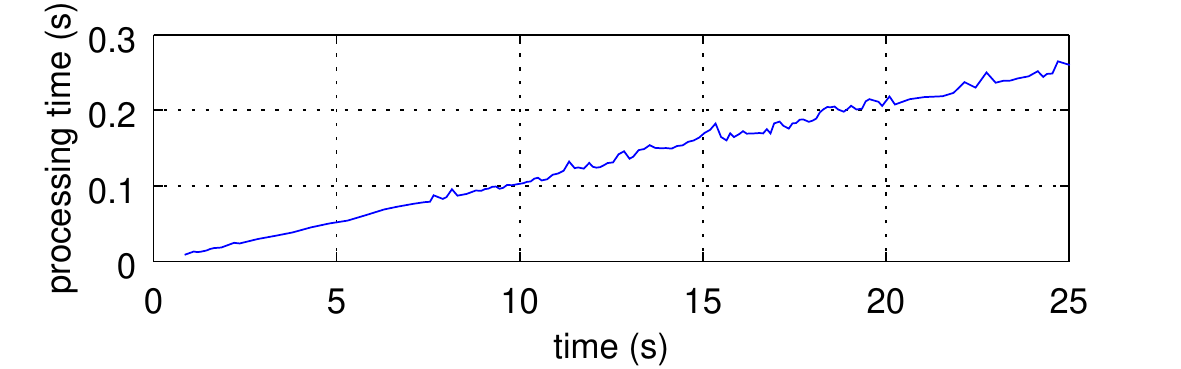}

      \caption{IMU initialization in \emph{V2\_01\_easy}.}
      \label{fig:ini}
   \end{figure}

\subsection{SLAM Evaluation and Comparison to State-of-the-Art}

\begin{table*}[t] 
\caption{Keyframe Trajectory Accuracy in EuRoC Dataset (Raw Ground-Truth)}\label{t:acc}
\begin{center}
\begin{tabular}{|l|c|c|c|c|c|c|c|c|}
\cline{2-9}
 \multicolumn{1}{c}{}&  \multicolumn{6}{|c|}{} & \multicolumn{2}{|c|}{}  \\[-0.9em]
 \multicolumn{1}{c}{}&  \multicolumn{6}{|c|}{Visual-Inertial ORB-SLAM} & \multicolumn{2}{|c|}{Monocular ORB-SLAM}   \\[0.1em]
\cline{2-9}
 \multicolumn{1}{c}{}&  \multicolumn{3}{|c|}{} &  \multicolumn{3}{|c|}{} & &  \\[-0.9em]
 \multicolumn{1}{c}{}&  \multicolumn{3}{|c|}{No Full BA} &  \multicolumn{3}{|c|}{Full BA}  &  No Full BA & Full BA   \\[0.1em]
\cline{2-9}
 \multicolumn{1}{c|}{} & & & & & & & & \\[-0.9em]
\multicolumn{1}{c|}{}  & RMSE (m) & Scale & RMSE(m) & RMSE (m) & Scale & RMSE (m) & RMSE(m) & RMSE(m) \\[0.1em]
\multicolumn{1}{c|}{}  &  & Error (\%) & \emph{GT scale}$^*$ &   & Error (\%) & \emph{GT scale}$^*$  & \emph{GT scale}$^*$  & \emph{GT scale}$^*$  \\[0.1em]
\cline{2-9}
\hline
 & & & & & & & & \\[-0.9em]
V1\_01\_easy & 0.027 & 0.9& 0.019 & 0.023& 0.8 & 0.016 &0.015& 0.015\\[0.1em]
\hline
 & & & & & & & & \\[-0.9em]
V1\_02\_medium & 0.028 & 0.8 & 0.024 & 0.027 & 1.0 & 0.019 & 0.020& 0.020\\[0.1em]
\hline
 & & & & & & & & \\[-0.9em]
V1\_03\_difficult & X & X & X & X & X & X & X & X  \\[0.1em]
\hline
\hline
 & & & & & & & & \\[-0.9em]
V2\_01\_easy & 0.032 & 0.2 & 0.031 &0.018& 0.2 & 0.017 &0.021 & 0.015\\[0.1em]
\hline
 & & & & & & & & \\[-0.9em]
V2\_02\_medium & 0.041 & 1.4 & 0.026 & 0.024 & 0.8 & 0.017 & 0.018 & 0.017\\[0.1em]
\hline
 & & & & & & & & \\[-0.9em]
V2\_03\_difficult & 0.074 & 0.7& 0.073& 0.047 & 0.6  & 0.045 & X & X \\[0.1em]
\hline
\hline
 & & & & & & & &\\[-0.9em]
MH\_01\_easy & 0.075 & 0.5& 0.072& 0.068& 0.3 & 0.068  &0.071 & 0.070\\[0.1em]
\hline
 & & & & & & & &\\[-0.9em]
MH\_02\_easy & 0.084& 0.8 & 0.078 &0.073 & 0.4 & 0.072 & 0.067 & 0.066 \\[0.1em]
\hline
 & & & & & & & &\\[-0.9em]
MH\_03\_medium & 0.087 & 1.5 & 0.067 &0.071 & 0.1 & 0.071 &0.071&  0.071\\[0.1em]
\hline
 & & & & & & & & \\[-0.9em]
MH\_04\_difficult & 0.217 & 3.4 & 0.081 &0.087 &  0.9 & 0.066 & 0.082 & 0.081\\[0.1em]
\hline
 & & & & & & & & \\[-0.9em]
MH\_05\_difficult & 0.082 & 0.5 & 0.077 & 0.060 &  0.2 & 0.060 &0.060 & 0.060 \\[0.1em]
\hline
\end{tabular} 
\end{center}
$^*$\emph{GT scale}: the estimated trajectory is scaled so that it perfectly matches the scale of the ground-truth. These columns are included for comparison purposes but do not 
represent the output of a real system, but the output of an \emph{ideal} system that could estimate the true scale.
\end{table*}

\newcommand{\scalec}{0.39}
\begin{figure*}[t!] 
  \centering
      \includegraphics[width=0.55\textwidth]{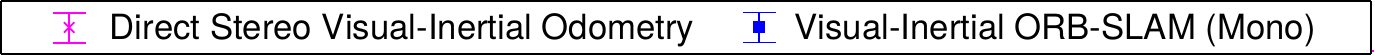}
      \\      
      \includegraphics[width=\scalec\textwidth]{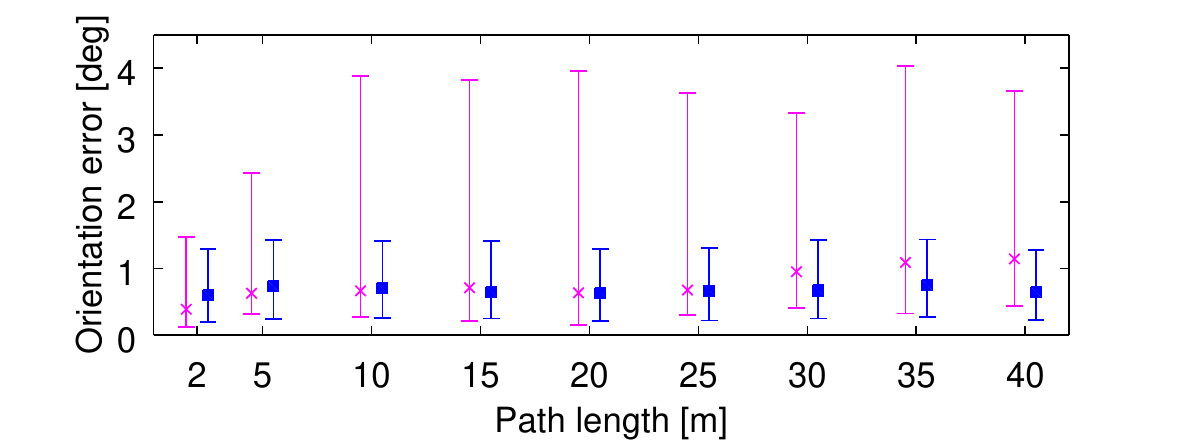}
      \quad
     \includegraphics[width=\scalec\textwidth]{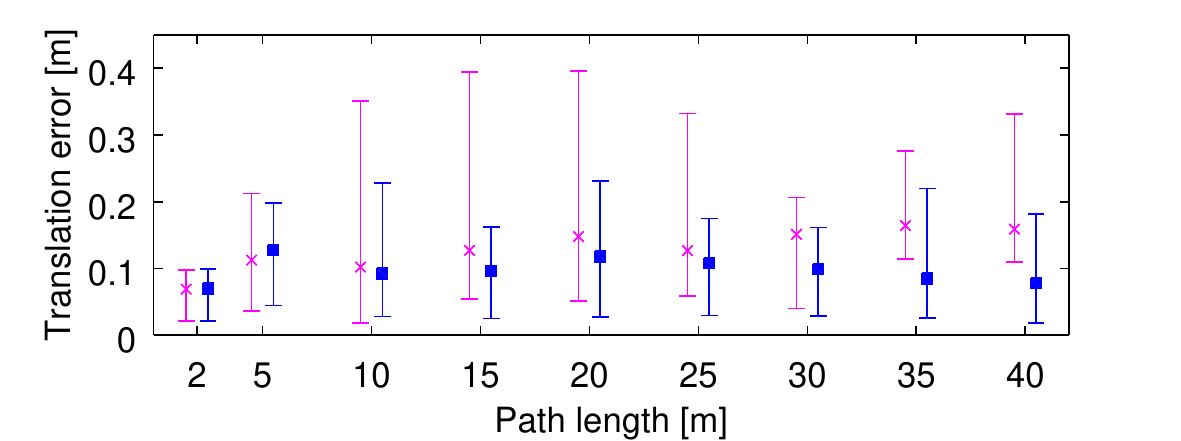}
      \\[-0.6em]
      \small Sequence: {V1\_01\_easy}
      \\[0.5em]
     \includegraphics[width=\scalec\textwidth]{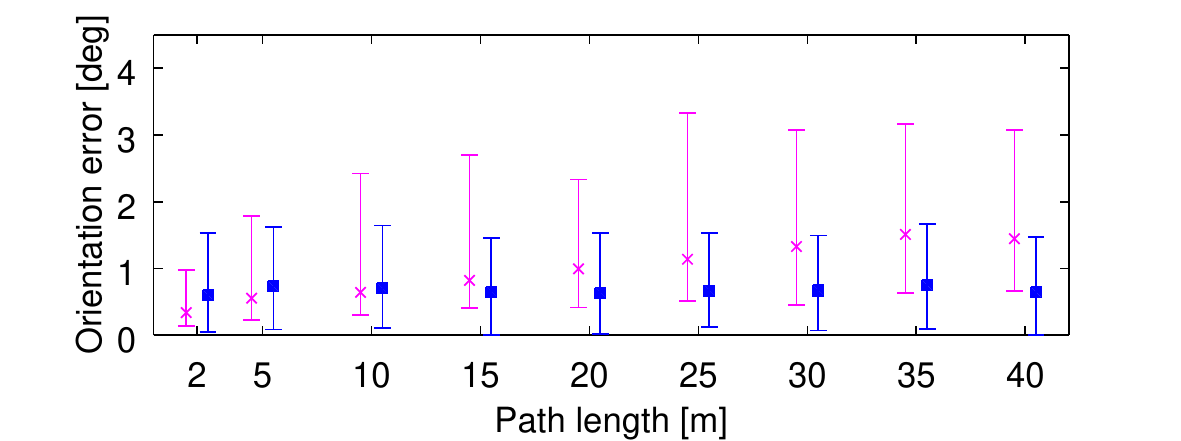}
     \quad
     \includegraphics[width=\scalec\textwidth]{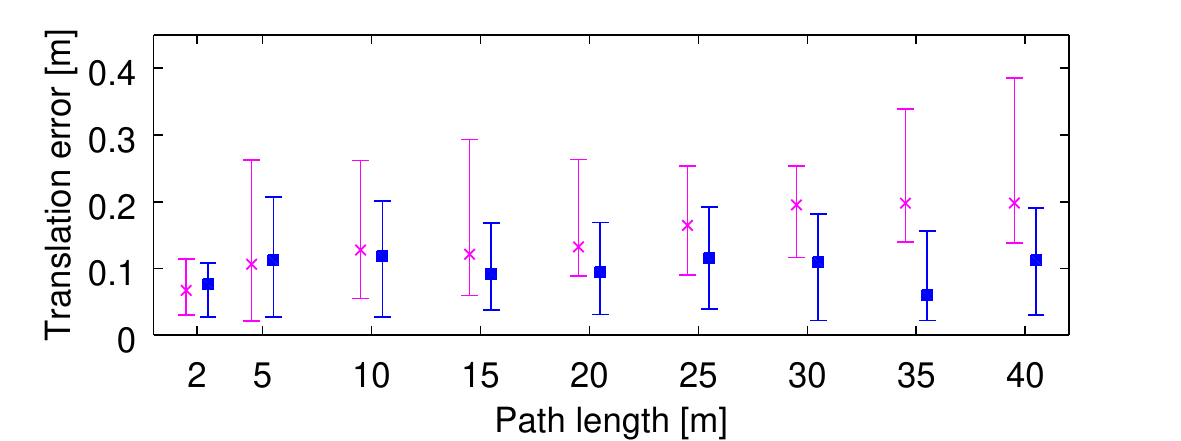}   
     \\[-0.6em]
      Sequence: {V1\_02\_medium}
      \caption{Relative Pose Error \cite{Geiger2012CVPR} comparison between our approach and the state-of-the-art direct stereo visual-inertial odometry \cite{EngelIMU}. The error for our SLAM system does not grow for longer paths, due
      to map reuse, in contrast to the visual-inertial odometry method where drift cannot be compensated. Note that \cite{EngelIMU} uses stereo, while our results are monocular.}
      \label{fig:comp}
   \end{figure*}

We evaluate the accuracy of our Visual-Inertial ORB-SLAM in the 11 sequences of the EuRoC dataset. We start processing the sequences when the MAV starts exploring.
The local window size for the local BA is set to 10 keyframes and
the IMU initialization is performed after 15 seconds from monocular ORB-SLAM initialization. The system performs a full BA just after IMU initialization. Table \ref{t:acc} shows the
translation Root Mean Square Error (RMSE) of the 
keyframe trajectory for each sequence, as proposed in \cite{tumrgbd}. We use the raw Vicon and Leica ground-truth as the post-processed one already used IMU. We observed a 
time offset 
between the visual-inertial sensor and the raw ground-truth of $-0.2s$ in the \emph{Vicon} \emph{Room} \emph{2} sequences and $0.2s$ in the \emph{Machine} \emph{Hall}, that 
we corrected when aligning both trajectories. We also measure the ideal scale factor that would align optimally the estimated trajectory and ground-truth. 
This scale factor can be regarded as the residual scale error of the trajectory and reconstruction. 
Our system successfully processes all these sequences in real-time, except \emph{V1\_03\_difficult}, 
where the movement is too extreme for the monocular system to survive 15 seconds. Our system is able to recover motion with 
metric scale, with a scale error typically below $1\%$, achieving a typical precision of $3\mathrm{cm}$ for $30\mathrm{m}^2$ room environments and 
of $8\mathrm{cm}$ for $300\mathrm{m}^2$ industrial environments. To show the loss in accuracy due to scale error, we also show the RMSE if the system
would be able to recover the true scale, see \emph{GT scale} columns. We also show that the precision and scale estimation 
can be further improved by performing a visual-inertial full BA at the end of the execution, see Full BA columns.
The reconstruction for sequence \emph{V1\_02\_medium} can be seen in Fig. \ref{fig:view}, and in the accompanying video\footnote{\url{https://youtu.be/rdR5OR8egGI}}.

To contextualize our results, we include as baseline the results of our vision-only system in Table \ref{t:acc}. Our visual-inertial system is more
robust as it can process \emph{V2\_03\_difficult}, it is able to recover metric scale and does not suffer scale drift. The accuracy of the visual-inertial system is
similar to the accuracy that would obtain the vision-only version if it could ideally recover the true scale. However the visual-inertial bundle adjustment is more 
costly, as explained in Section \ref{sec:mapping}, and the local window of the local BA has to be smaller that in the vision-only case. 
This explains the slightly worse results of the \emph{GT scaled} visual-inertial results
without full BA. In fact the visual-inertial full BA typically converges in 15 iterations in 7 seconds, while the vision-only full BA converges in 5 iterations in less than 1 second.

In order to test the capability of Visual-Inertial ORB-SLAM to reuse a previous map, we run in a row all sequences of the same environment. We process the first sequence and perform
a full BA. Then we run the rest of the sequences, where our system relocalizes and continue doing SLAM. We then compare the accumulated keyframe trajectory with the 
ground-truth. As there exists a previous map, our system is now able to localize the camera in sequence \emph{V1\_03\_difficult}. 
The RMSE in meters for V1, V2 and MH environments are 0.037, 0.027 and 0.076 respectively, with an scale factor error of 1.2\%, 0.1\% and 0.2\%. A final full BA has a negligible 
effect as we have already performed a full BA at the end of the first sequence. These results show that there is no drift accumulation when revisiting the same scene, as the 
RMSE for all sequences is not larger than for individual sequences. 

Finally we have compared Visual-Inertial ORB-SLAM to the state-of-the-art direct visual-inertial odometry for stereo cameras \cite{EngelIMU}, 
which also showed results in \emph{Vicon Room 1} sequences,
 allowing for a direct comparison. Fig. \ref{fig:comp} shows the Relative Pose Error (RPE) \cite{Geiger2012CVPR}. To compute the RPE for our method, we need to recover the frame 
 trajectory, as only keyframes are optimized by our backend. To this end, when tracking a frame we store a relative transformation to a reference keyframe, so that we 
 can retrieve the frame pose from the estimated keyframe pose at the end of the execution. We have not run a full BA at the end of the experiment. We can see that the error for the 
 visual-inertial odometry method grows with the traveled distance, 
 while our visual-inertial SLAM system does not accumulate error due to map reuse. The stereo method \cite{EngelIMU} is able to work in \emph{V1\_03\_difficult}, while our monocular method fails.
 Our monocular SLAM successfully recovers metric scale, and achieves comparable accuracy in short paths, where the advantage of SLAM is negligible compared to odometry. 
 This is a remarkable result of our feature-based monocular method, compared to \cite{EngelIMU} which is direct and stereo.

\section{CONCLUSIONS}
We have presented a novel tightly coupled Visual-Inertial SLAM system, that is able to close loops in real-time and localize the sensor reusing the map in already mapped areas.
This allows to achieve a  \emph{zero-drift}  localization, in contrast to visual odometry approaches where drift grows unbounded. The experiments show that our monocular SLAM 
recovers metric scale with high precision, and achieves better accuracy than the state-of-the-art in stereo visual-inertial odometry when continually 
localizing in the same environment. We consider this \emph{zero-drift} localization of particular interest for virtual/augmented reality systems, where the predicted user viewpoint
must not drift when the user operates in the same workspace. Moreover we expect to achieve better accuracy and robustness by using stereo or RGB-D cameras, which
would also simplify IMU initialization as scale is known.
The main weakness of our proposed IMU initialization is that it relies on the initialization of the monocular SLAM. We would like to investigate the use of the gyroscope to
make the monocular initialization faster and more robust.

\section*{ACKNOWLEDGMENT}
We thank the authors of \cite{EUROC} for releasing the EuRoC dataset and the authors of \cite{EngelIMU} for providing their data for the comparison in Fig. \ref{fig:comp}.

\end{document}